\documentclass[runningheads]{llncs}
\usepackage{graphicx}
\usepackage{xcolor}
\usepackage{float}
\usepackage{adjustbox}
\usepackage{blindtext}
\usepackage{hyperref}

\begin{document}

\newcommand{\nick}[1]{\textcolor{red}{Nick : #1}}
\newcommand{\ziyan}[1]{\textcolor{blue}{Ziyan : #1}}

\title{EduSAT: A Pedagogical Tool for Theory and Applications of Boolean Satisfiability}
\titlerunning{EduSAT}
%
\author{Yiqi Zhao\inst{1, 2}\orcidID{0009-0007-4283-6358} \and
Ziyan An\inst{1}\orcidID{0000-0002-1083-0011} \and \\
Meiyi Ma\inst{1}\orcidID{0000-0001-6916-8774} \and Taylor Johnson\inst{1}\orcidID{0000-0001-8021-9923}}
\authorrunning{Y. Zhao et al.}
%
\institute{Vanderbilt University, Nashville TN 37235, USA \email{taylor.johnson@vanderbilt.edu}\and
University of Southern California, Los Angeles, CA 90089, USA}
\maketitle              
\begin{abstract}
Boolean Satisfiability (SAT) and Satisfiability Modulo Theories (SMT) are widely used in automated verification, but there is a lack of interactive tools designed for educational purposes in this field. To address this gap, we present EduSAT, a pedagogical tool specifically developed to support learning and understanding of SAT and SMT solving.
EduSAT offers implementations of key algorithms such as the Davis-Putnam-Logemann-Loveland (DPLL) algorithm and the Reduced Order Binary Decision Diagram (ROBDD) for SAT solving. Additionally, EduSAT provides solver abstractions for five NP-complete problems beyond SAT and SMT.
Users can benefit from EduSAT by experimenting, analyzing, and validating their understanding of SAT and SMT solving techniques. Our tool is accompanied by comprehensive documentation and tutorials, extensive testing, and practical features such as a natural language interface and SAT and SMT formula generators, which also serve as a valuable opportunity for learners to deepen their understanding.
Our evaluation of EduSAT demonstrates its high accuracy, achieving 100\% correctness across all the implemented SAT and SMT solvers. We release EduSAT as a python package in .whl file, and the source can be identified at \href{https://github.com/zhaoy37/SAT\_Solver}{https://github.com/zhaoy37/SAT\_Solver}.

\end{abstract}

\section{Introduction}
Formal verification is a critical field with a focus to prove or disprove the safety, liveness, and other properties of a system using mathematical techniques such as induction. It is widely applied to circuit design \cite{b1}, autonomous vehicles \cite{b2}, and operating systems \cite{b3}, etc.  Many solutions to verification problems relate to the solutions of Boolean Satisfiability Problems (SAT) and its variant, Satisfiability Modulo Theories (SMT). Bounded Model Checking (BMC) of a discrete system, for instance, can be encoded as an equivalent SAT problem and solved with a domain specific solver such as Z3 \cite{b4}. Furthermore, by the Cook-Levin Theorem, SAT is NP-complete, which suggests that problems that are in NP, such as graph coloring, can be reduced to a SAT problem by a deterministic Turing Machine in polynomial time.

Despite the significance of SAT and SMT, we observe a lack of open-source solvers for SAT problems with a pedagogical purpose not to mention demonstrations of how some classical NP-complete problems are encoded into SAT or SMT formulations. One of the most commonly used SMT solver is Z3 \cite{b4}, which is based on DPLL and incorporates quantifier instatiation using E-matching and relevancy propagation, etc. Despite that the codes from Z3 are available for public use, Z3 is designed for efficient solutions in real-life circumstances without a focus on the educational value. Other state-of-the-art SMT solvers include CVC4 (Cooperating Validity Checker) \cite{b5}, which is also based on DPLL(T) and uses simplex method for its arithmetic, and Yices 2 \cite{b6}, which uses a SAT solver with CDCL approach. LearnSAT \cite{b7}, in comparison, is an educational tool that introduces techniques for SAT solving through the language of prolog. Despite the conciseness of logic programming, most students undergoing Computer Science educations are more exposed to and most oftenly use procedural and object-oriented programming paradigms. To bridge this gap and guide users through the object-oriented designs, we introduce EduSAT, a python-based Boolean Satisfiability Solver with support for SMT over integer predicates, corresponding formula generators and natural language interfaces, as well as extensions for solvers of 5 NP-complete problems (apart from SMT and SAT). We also introduce an interactive tutorial built in jupyter notebook and documentations in github.

\begin{figure}[hbt!]
\centering
\includegraphics[width=9cm]{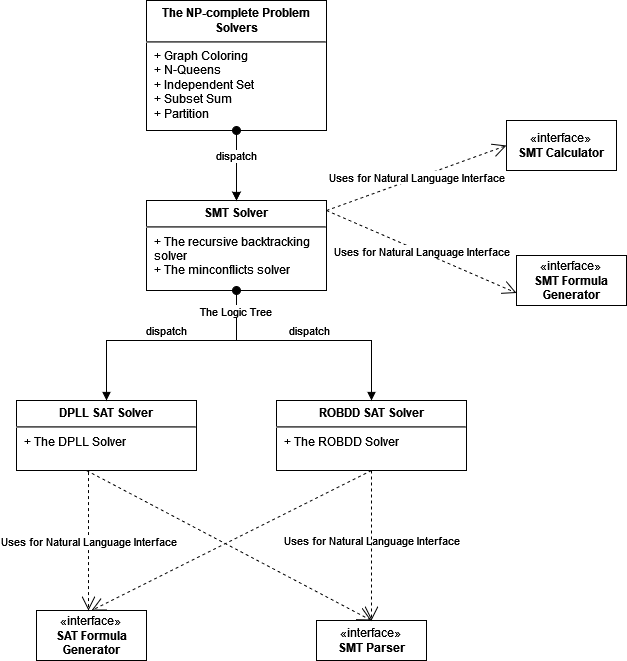}
\caption{Architecture of EduSAT}
\label{fig:architecture}
\end{figure}

We present the architecture of our educational tool in Figure \ref{fig:architecture}.

\section{Tool Details}

SAT refers to the problem to determine if there exists a model to a Boolean formula $P$ such that the model evaluates to true. SMT, on the other hand, extends the acceptable formula, $P$ to many-sorted formulas. In EduSAT, we only consider formulas with integer variables with the following operators: $+, -, \times, //, >, <, \le, \ge, =$, where $//$ represents integer division. For SMT problems, we also limit our solvers to a finite search space specified by the user and guarantee correctness only in this range, since the focus is in introducing the solving techniques.

\subsection{SAT Solving}
We implement two SAT solvers, the DPLL solver and the ROBDD solver. The DPLL solver uses recursive backtracking with heuristics revolving around assignment decidings and deducings. We focus on 3 heuristics specifically: Early termination refers to terminating the algorithm if any realization of an atom leads to True or False of a SAT formula in the search process. Unit clauses refers to the simplification of the SAT formula left with one literal via realizing that literal. Pure literals refers to that if all appearances of a symbol in a negation normal form of the SAT formula have the same sign across the clause, we can guess that the symbol is True (for positive symbol) or False (for negative symbol) in assigning during recursive backtracking.

The ROBDD solver provides an ordered canonical form of logic representation with a directed acyclic graph that consists of multiple decision nodes, two terminal nodes, and the corresponding edges. The terminal nodes in a ROBDD represent the constants true $\top$ and false $\bot$, the decision nodes represent variables or parameters, and the edges $E \in \{low, high\}$ represent the assignment of truth values to variables, where the low edges represent the assignment of false, and the high edges represent the assignment of true. The construction of a ROBDD from a Binary Decision Tree follows the elimination rule and isomorphic rule~\cite{b8}, such that there are no redundant nodes with identical sub-trees, and all terminal nodes are combined. 

\subsection{SMT Solving and NP-complete Problems}
In SMT solving, we first encode each arithmetic clause appeared in an SMT problem as an atom of a SAT problem, which is solved accordingly using the aforementioned SAT solvers. We then substitute all possible realizations of the SAT problem to construct a Constraint Satisfaction Problem (CSP) equivalent to the SMT formula and allow it to be solved through 2 methods: recursive backtracking and min-conflicts. In recursive backtracking, the solver performs an exhaustive search for the valuations of the SMT variables over a user-specified range. In min-conflicts, the algorithm first randomly initializes the values for the set of all SMT variables. While the assignment does not satisfy the formula, the algorithm randomly picks a variable from all the SMT variables that result in conflict(s), and find the value that best reduces the conflicts associated with that variable to update the assignment. The algorithm terminates when an assignment is found to satisfy the SMT formula or when the maximum number of iteration steps is achieved. Please note that the algorithm may report unsatisfiability even if there exists a solution if the maximum iteration steps is not sufficiently large (which rarely occurs for a large number of iterations). However, this approach can be faster than the naive backtracking approach when applied to some of our NP-complete problem solver(s) such as the N-queens problem solver.

With the implemented SMT solvers, we create NP-complete problem solver abstractions for 5 problems listed in Figure \ref{fig:architecture}. For example, in the N-queens problem, given an n by n sized chess board, the algorithm is asked to place n queens on the board such that no two queens attack each other. We encode this as an equivalent SMT problem: Each SMT variable represents a column, and the assignmnent to that variable represents the row index of the queen in that column. Then, in the SMT clause, the solver is supported with representations that no two queens can be in the same row and no two queens can be in the same diagonal. We solve the SMT problem accordingly to find the solution to the N-queens problem.

\section{Evaluations}

In this section, we demonstrate the accuracy and the efficiency of the SAT solvers. Due to the space limitation, we leave other evaluations (on the SMT solver for instance) in the GitHub repository.

\subsection{Evaluation on the SAT Solvers}
\label{eval:sat}
\subsubsection{DPLL Solver}
All evaluations for the DPLL Solver were conducted on an Apple M1 Pro chip with 16 GB memory. In terms of efficiency, we evaluated how the effects of the number of formulas solved and the difference between single vs. multiple solutions affected the speed of the DPLL solver through controlled experiment trials. In each experimental trial, we generated random Boolean logic trees (with chance of not node, and node and or node held constant at 0.1, 0.45, and 0.45 respectively) and provided solutions to such trees using both a naive tabular solution (recursive backtracking without heuristic) and the DPLL solver with heuristics and compared the performances.

In terms of the accuracy, we evaluated the solutions provided by both the naive solver and the solver with heuristics directly using the evaluation feature of the logic trees instead of formula simplifications performed by the solver. For the cases of unsatisfiability and the completeness of solutions (when it comes to inquiries of multiple solutions), we cross checked the results generated by the DPLL solver and by the naive recursive backtracking solver.

We observe \textbf{100\% accuracy} for both of the naive solver and the solver with heuristics for all trials we performed. Moreover, the solver with heuristics outperform the naive solver in terms of efficiency in all trials we performed. We demonstrate the evaluation results (in terms of efficiency) in Table \ref{table:dpll_eval_results}.
All time units are in seconds.

\begin{table}[t]
\begin{center}
\caption{\label{table:dpll_eval_results}Efficiency of the DPLL Solver with Varying Number of Formulas}
\begin{adjustbox}{max width=\textwidth}
\begin{tabular}{||c |c |c |c |c |c ||} 
 \hline
 No. Formulas & Single or not & No. Variables & Depth & Solving Time (No Heuristic) & Solving Time (Heuristic)\\
 \hline
 10 & Single & 5 & 8 & 0.0064& 0.0016\\
 \hline
 100 & Single & 5 & 8 & 0.072& 0.014\\
 \hline
 1000 & Single & 5 & 8 & 0.76& 0.15\\
 \hline
 10 & Multiple & 5 & 8 & 0.025& 0.0032\\
 \hline
 100 & Multiple & 5 & 8 & 0.17& 0.034\\
 \hline
 1000 & Multiple & 5 & 8 & 1.74& 0.35\\
 \hline
\end{tabular}
\end{adjustbox}
\end{center}
\end{table}

\subsubsection{ROBDD Solver}
All experiments were performed on an Intel Core i9-10850K CPU running at 3.60GHz. The accuracy and efficiency of the ROBDD SAT solver were evaluated using the same format as DPLL. To measure efficiency, we recorded the runtime from the conversion of a BDT to a ROBDD to the completion of finding all SAT solutions. To evaluate accuracy, we generated different numbers of logic formulas with varying numbers of variables and depths, then verified the solutions provided by the ROBDD solver by substituting them back into the formulas. Table~\ref{table:rodbb_varying_num_formulas} presents the evaluation results. 

\begin{table}[!h]
\begin{center}
\caption{\label{table:rodbb_varying_num_formulas}ROBDD solver with varying number of formulas, variables, and depth.}
\begin{adjustbox}{max width=\textwidth}
\begin{tabular}{||c |c |c |c |c |c |c ||} 
 \hline
 Formulas & Parameters & Depth & Single Solution (s) & Multiple Solutions (s) & Accuracy (Single) & Accuracy (Multiple) \\
 \hline
 10 & 5 & 5 & 0.0027 & 0.0030 & 100\% & 100\%\\
 \hline
 10 & 5 & 6 & 0.0027 & 0.0033 & 100\% & 100\%\\
 \hline
 10 & 5 & 7 & 0.0028 & 0.0035 & 100\% & 100\%\\
 \hline
 10 & 3 & 3 & 0.0011 & 0.0009 & 100\% & 100\% \\
 \hline
 100 & 3 & 3 & 0.0090 & 0.0100 & 100\% & 100\%\\
 \hline
 1000 & 3 & 3 & 0.1000 & 0.1075 & 100\% & 100\%\\
 \hline
\end{tabular}
\end{adjustbox}
\end{center}
\end{table}

Note that for multiple solutions tasks, we find all possible solutions to a logic formula, while for single solution tasks, we always provide the \textit{shortest and simplest} solutions. 
We observe a slight increase in runtime for multiple solution tasks compared to single solution tasks. 
It is also evident that the total solving time generally increases as the number of formulas and the depth increase, which is as expected. 
We note that the ROBDD solver achieved \textbf{100\% accuracy} in both the single solution task and the multiple solution task.

\section{An Illustrative Case Study on ROBDD}

\begin{figure}[!h]
\centering
  \begin{minipage}[t]{0.45\textwidth}
    \centering
    \includegraphics[width=\linewidth]{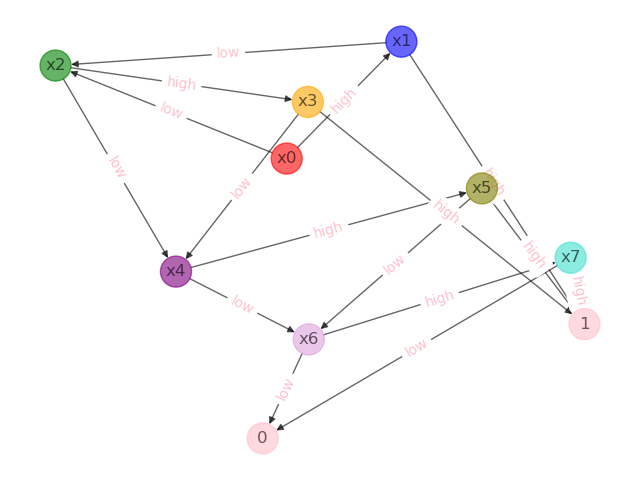}
    \caption{Variable order: $x_0, \ldots x_7$.}
    \label{fig:image3}
  \end{minipage}
  \begin{minipage}[t]{0.45\textwidth}
    \centering
    \includegraphics[width=\linewidth]{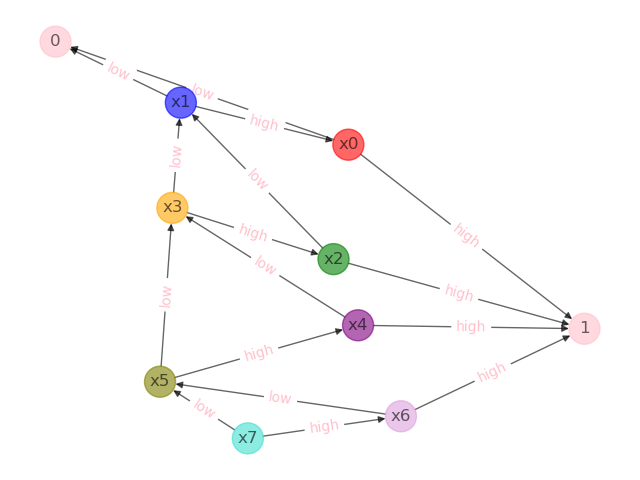}
    \caption{Variable order: $x_7, \ldots x_0$.}
    \label{fig:image4}
  \end{minipage}
\end{figure}

We present visual demonstrations of the ROBDD functionality in Fig.\ref{fig:image3} and Fig.\ref{fig:image4}. Specifically, they showcase the ROBDD with different parameter orders for a complex formula, $f(x_0, x_1, x_2, x_3, x_4, x_5, x_6, x_7)= ((x_0 \wedge x_1) \vee (x_2 \wedge x_3) \vee (x_4 \wedge x_5) \vee (x_6 \wedge x_7))$. Although our tool offers truth table visualizations, we find that truth tables become significantly less efficient as the number of parameters increases.
Overall, the visualization of ROBDDs provides a clear and concise understanding of the logic formula structure, making it easier for users to identify potential issues with their logic.

\section{Conclusion}
We construct a pedagogical tool, EduSAT, for SAT solving and its applications. We introduce essential concepts such as DPLL, ROBDD, recursive backtracking in SMT solving, min-conflicts, as well as methodologies for solving NP-complete problems via SMT formulations. The tool achieves a high accuracy and efficiency. At the same time, we implement detailed documentations, tutorials, and interactive components such as the natural language interface and the ROBDD visual demonstrations to allow better understanding from the users.

\end{document}